\documentclass{article}

\usepackage[preprint]{neurips_2024}

\usepackage[utf8]{inputenc} 
\usepackage[T1]{fontenc}    
\usepackage{hyperref}       
\usepackage{url}            
\usepackage{booktabs}       
\usepackage{amsfonts}       
\usepackage{nicefrac}       
\usepackage{microtype}      
\usepackage{xcolor}         
\usepackage{amsmath}
\usepackage{amssymb}
\usepackage{nomencl}
\usepackage{glossaries}
\usepackage{xspace}
\usepackage{multirow}
\usepackage[pdftex]{graphicx}
\usepackage{bbding}
\usepackage{pifont}
\usepackage{algorithm}
\usepackage{subfigure}
\usepackage{enumitem}
\usepackage{multicol}
\usepackage{caption}
\usepackage{array}
\usepackage{fp}
\usepackage{makecell}
\usepackage{flushend}
\usepackage{cases}
\usepackage{color}
\usepackage{algpseudocode}
\usepackage{listings}
\usepackage{mathrsfs}
\usepackage{longtable}
\usepackage{colortbl}
\usepackage{bm}

\definecolor{codegreen}{rgb}{0,0.3,0.6}
\definecolor{codegray}{rgb}{0.5,0.5,0.5}

\newcommand{\ie}{\emph{i.e.,}\xspace}
\newcommand{\eg}{\emph{e.g.,}\xspace}

\newcommand{\ignore}[1]{}

\usepackage{tcolorbox}
\definecolor{darkorange}{RGB}{255, 140, 0}
\definecolor{lightgreen}{RGB}{145, 204, 117}
\definecolor{lightyellow}{RGB}{250, 200, 88}
\definecolor{lightred}{RGB}{238, 102, 102}
\definecolor{lightblue}{RGB}{115, 192, 222}
\newtcolorbox{promptbox}[2][Prompt]{
colback=black!5!white,
arc=5pt, 
boxrule=0.5pt,
fonttitle=\bfseries,
title=#1, 
before upper={\scriptsize}, fontupper=\fontfamily{ptm}\selectfont,
colframe=#2, 
}

\title{Virgo: A Preliminary Exploration on \\ Reproducing o1-like MLLM}

\author{%
    Yifan Du$^{1,2}$\thanks{Equal contribution.}~,
  Zikang Liu$^{1*}$,
  Yifan Li$^{1*}$,\\
  \textbf{Wayne Xin Zhao$^{1}$\thanks{Correspondence to Wayne Xin Zhao.}~, 
  Yuqi Huo$^2$, Bingning Wang$^2$, Weipeng Chen$^2$} \\
  \textbf{Zheng Liu}$^3$, \textbf{Zhongyuan Wang}$^3$, \textbf{Ji-Rong Wen}$^{1}$
  \\
  $^1$Gaoling School of Artificial Intelligence, Renmin University of China.\\
  $^2$Baichuan AI.\\
  $^3$BAAI.\\
  \texttt{\{yifandu1999, jasonlaw8121, liyifan0925, batmanfly\}@gmail.com}\\
}

\begin{document}
Technical Report on Slow Thinking with LLMs: Visual Reasoning

\maketitle

\begin{figure}[ht]
    \centering    \includegraphics[width=0.6\textwidth]{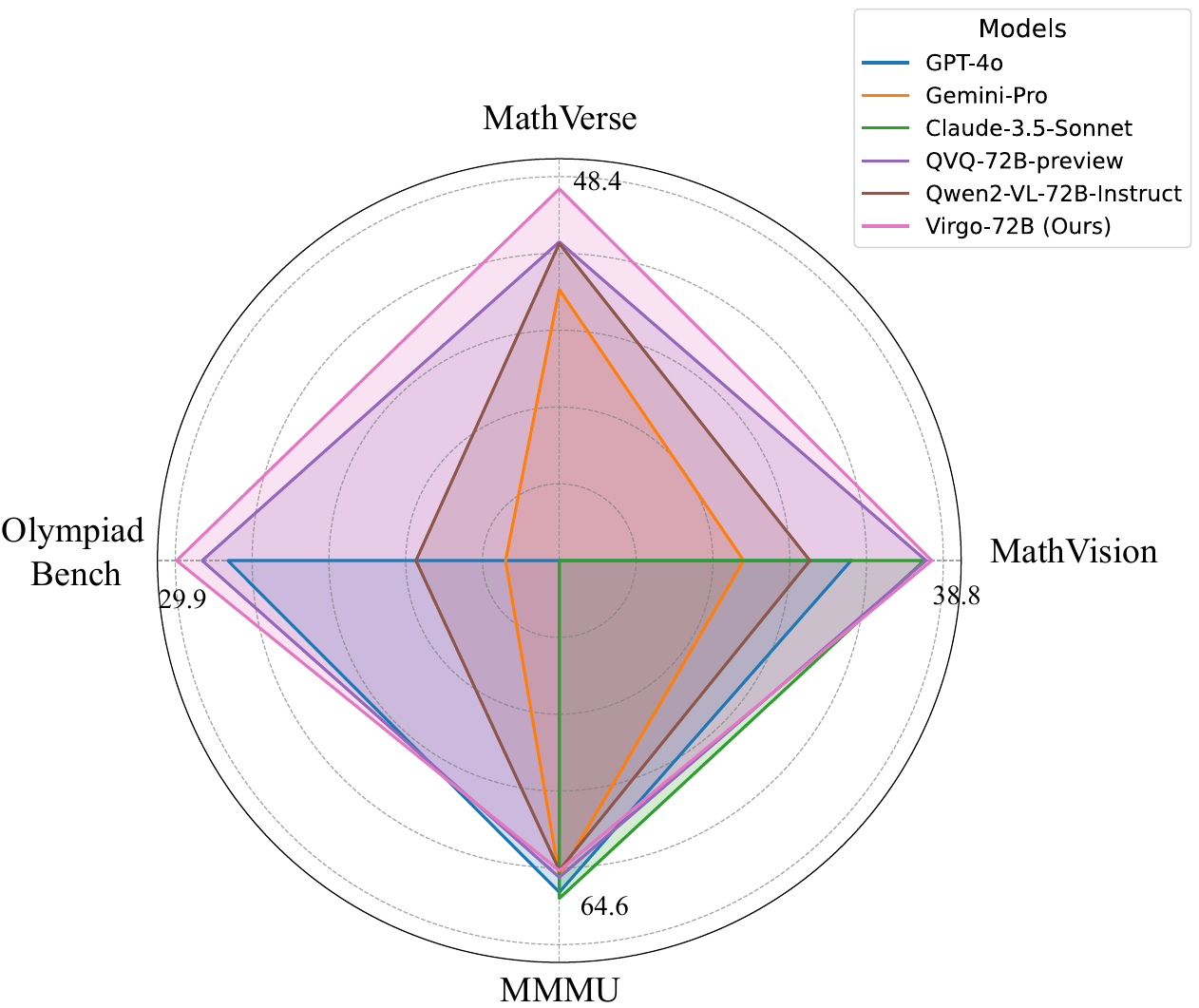}
  \caption{The radar chart compares the performance of top-tier MLLMs across four challenging multimodal benchmarks. Our model, Vigor-72B (fine-tuned from Qwen2-VL-72B-Instruct),  demonstrates leading performance.}
  \label{fig:radar} 
\end{figure}
\begin{abstract}
Recently, slow-thinking reasoning systems, built upon large language models (LLMs), have garnered widespread attention by scaling the thinking time during inference. There is also growing interest in adapting this capability to multimodal large language models (MLLMs). Given that MLLMs handle more complex data semantics across different modalities, it is intuitively more challenging to implement multimodal slow-thinking systems. 

To address this issue, in this paper, we explore a straightforward approach by fine-tuning a capable MLLM with a small amount of textual long-form thought data, resulting in a multimodal slow-thinking system, \textbf{Virgo} (\textbf{Vi}sual \textbf{r}easoning with lon\textbf{g} th\textbf{o}ught). We find that these long-form reasoning processes, expressed in natural language, can be effectively transferred to MLLMs. Moreover, it seems that such textual reasoning data can be even more effective than visual reasoning data in eliciting the slow-thinking capacities of MLLMs. While this work is preliminary, it demonstrates that slow-thinking capacities are fundamentally associated with the language model component, which can be transferred across modalities or domains. This finding can be leveraged to guide the development of more powerful slow-thinking reasoning systems. We release our resources at \url{https://github.com/RUCAIBox/Virgo}.
\end{abstract}

\section{Introduction}
Recently, slow-thinking reasoning systems (\eg  OpenAI o1~\cite{o1}, DeepSeek R1~\cite{r1}, and Qwen QwQ~\cite{qwq}) have demonstrated significant performance improvements across various benchmarks, particularly excelling in challenging problems where previous large language models (LLMs)~\cite{zhao2023survey} have underperformed~\cite{wu2024comparative,zhong2024evaluation,openai2024reasoning}. These systems employ both train-time and test-time scaling to enhance accuracy and capacity in solving complex tasks, which typically involve an extended reasoning process, referred to as ``\emph{thought}'', before reaching the final solution. While these systems primarily focus on textual problems, the latest advancements have also shown promising results in multimodal scenarios (\eg QVQ~\cite{qvq}).

Although commercial companies have not disclosed the underlying techniques for creating slow-thinking systems, researchers have made significant efforts to advance public technical progress in this area. According to the existing literature, there are two typical approaches to implementing slow-thinking reasoning systems~\cite{jiang2024technical,zhang2024llamaberry,zhang2024o1coder,qin2024o1replicationjourneystrategic,zhao2024marco}. The first approach utilizes an explicit search structure (\eg Monte Carlo tree search) and employs specially trained reward models to guide the search process toward the correct path. The second approach involves distilling or constructing instruction data in the form of long chain-of-thought (CoT) and then fine-tuning a capable LLM to learn such reasoning modes. This method can be further enhanced using self-improvement or self-play methods such as direct preference optimization and reinforcement learning.


A major limitation of these research studies is their primary focus on textual tasks, with relatively little consideration given to multi-modal scenarios. To our knowledge, existing efforts in developing multimodal reasoning systems still significantly lag behind commercial systems like o1 and QVQ. This paper aims to explore the implementation of multimodal slow-thinking reasoning systems that can achieve performance comparable to these commercial systems. Unlike text-based reasoning systems, multimodal reasoning systems (\ie Multimodal LLMs, or MLLMs)~(\cite{li2024multimodal,yang2023mmreact}) often leverage both the abilities of perception (\ie understanding the visual semantics of the input figures) and reasoning (\ie determining the approach to solving a given task). Consequently, it is still unclear how slow-thinking operates on multimodal inputs.

Overall, we believe that developing multimodal slow-thinking reasoning systems presents a significant challenge, as it involves addressing the complex data and modeling mechanisms inherent in multi-modal tasks. 
Considering this challenge, we aim to explore a simple idea to implement multimodal reasoning systems:  
Can we directly adapt MLLMs by fine-tuning them with text-based long thought data? The intuition is that, since reasoning is primarily handled by the LLM component within an MLLM, we might be able to elicit its slow-thinking capacities using text-only instructions. In fact, there is growing evidence in enhancing the capabilities of MLLMs by leveraging more extensive text-only instruction~\cite{li2024temporal,zhang2024long}.  

In this paper, we design a straightforward approach to enhance MLLMs with slow-thinking capacities by leveraging text-based long thought data. Our primary focus is investigating two key questions: (1) Can the slow-thinking ability transfer across modalities through fine-tuning with text-based long thought data? (2) Can the ability derived from text-based long thought data be comparable to that distilled from multimodal slow-thinking systems? Specifically, we consider collecting textual reasoning instructions shared by previous studies~\cite{min2024imitate}, as well as generating visual reasoning instructions by distilling from multimodal slow-thinking systems. Following this method, we implement a multimodal slow-thinking systems, denoted as  \textbf{Virgo} (\textbf{Vi}sual \textbf{r}eaoning with lon\textbf{g} th\textbf{o}ught).
We select a capable MLLM, Qwen2-VL-72B-Instruct, as the backbone model and employ different instruction datasets to fine-tune this MLLM. This approach allows us to examine the reasoning performance of MLLMs when trained with various instructional datasets.




To study the research questions we propose, we conduct extensive experiments on four challenging benchmarks: MathVerse, MathVision, OlympiadBench, and MMMU. Our empirical study reveals that even when utilizing only text-based long thought data, Virgo achieves very promising results, comparable to or even surpassing those of commercial reasoning systems. Another major finding is that text-based reasoning instructions are generally more effective than multimodal reasoning data for eliciting the slow-thinking capacities of MLLMs. 

\section{Method}

In this section, we present our preliminary attempts to adapt MLLMs by equipping them with slow-thinking capacities for complex multimodal tasks. We explore two straightforward adaptation methods: (1) transferring slow-thinking abilities using text-based long thought data, and (2) distilling multimodal long thought data from existing slow-thinking MLLMs. Our aim is to investigate how slow-thinking capacities are elicited in MLLMs and to identify which approaches are more effective for achieving this goal. Next, we describe the specific implementation details.

\subsection{Capacity Transfer from Text-only Instructions}
\label{sec:text-transfer}
Previous studies~\cite{min2024imitate} have shown that slow-thinking reasoning is likely a behavioral mode that can be elicited by fine-tuning with a small amount of long thought data. Moreover, this capacity can generalize across different domains. Therefore, our idea is to investigate whether this ability can also transfer to different modalities, given that existing MLLMs are developed with LLMs as their backbone.


\subsubsection{Collecting Textual Long Thought Data}

We begin by collecting textual long thought data from our previous study~\cite{min2024imitate}. Specifically, we obtain approximately 5K long thought instruction instances distilled from two open slow-thinking reasoning systems: \texttt{DeepSeek-R1-Lite-Preview}~\cite{r1} (abbreviated as R1) and \texttt{QwQ-32B-preview}~\cite{qwq} (abbreviated as QwQ). The statistics of the collected instruction data are categorized by domain as follows: math (3.7K), science (0.9K), code (0.2K) and puzzle (0.1K). We select the majority of the instruction data from the math domain because it contains more challenging problems that require longer reasoning processes.

These instructional data are formatted with two distinct parts: the thought process, indicated by special symbols ``\texttt{<|begin\_of\_thought|>}'' and ``\texttt{<|end\_of\_thought|>}'', and the final solution, indicated by special symbols ``\texttt{<|begin\_of\_solution|>}'' and ``\texttt{<|end\_of\_solution|>}''. More details about the data composition and instruction format can be found in our previous paper~\cite{min2024imitate}.

\subsubsection{Textual Long Thought Instruction Tuning}
After collecting instruction data for long-form reasoning, we fine-tune the base MLLM to emulate slow-thinking reasoning behavior. We choose Qwen2-VL-72B-Instruct~\cite{wang2024qwen2} as the target model due to its excellent multimodal capabilities. Additionally, our previous work~\cite{min2024imitate} indicates that slow-thinking capacities are more readily achieved in stronger models.


To optimize the target MLLM, we train only the parameters from the LLM and cross-modal connector while keeping the parameters in the visual encoder frozen. We use the following optimization settings: a learning rate of 7e-6, a batch size of 128, and training for 10 epochs. Based on the performance on the development set, we select the model at the 5th epoch for evaluation.

We do not employ more advanced training algorithms, such as DPO~\cite{rafailov2024direct} and RLHF~\cite{Ouyang2022instruct}, as our objective is not to attain the maximum possible performance. Instead, we aim to explore the potential of transferring slow-thinking capacities through simple fine-tuning with textual long thought data. Our aim is to investigate the effect of textual long thought data using the straightforward imitation method.

\subsection{Capacity Distillation from  Slow-thinking MLLMs}


The second approach we explore is the direct distillation of multimodal long thought data from slow-thinking MLLMs (\eg QVQ). This approach aims to achieve two goals: first, to compare the fine-tuning performance of textual long thought data with that of multimodal long thought data, and second, to investigate the potential effects of combining both textual and multimodal instruction data.



\subsubsection{Visual Long Thought Data Collection}

To construct visual long thought data, a crucial step is to gather a set of high-quality visual problems, which include both task descriptions and images as input. Additionally, these problems should be accompanied by ground-truth answers for correctness verification. We consider selecting problems from visual question answering (VQA) datasets to cover diverse domains such as geometry, tables, figures, and icons. We select these domains because they typically present more challenging problems for MLLMs. 


Specifically, we select four geometry datasets (Geos~\cite{seo2015solving}, GeoQA+~\cite{chen2021geoqa}, Geometry3K~\cite{lu2021inter}, and UniGeo~\cite{chen2022unigeo}), three table and figure datasets (TabMWP~\cite{lu2022dynamic}, FigureQA~\cite{kahou2017figureqa}, and ChartQA~\cite{masry2022chartqa}), and an object dataset (CLEVR~\cite{johnson2017clevr}). These datasets can be accessed from the LLaVA-OneVision~\cite{li2024llava} data collection, where each instance provides a question, image, and answer triple. Detailed statistics for each dataset are presented in Table~\ref{tab:data_summarize}.

\begin{table}[ht]
    \caption{The statistics of the selected visual query sources}
    \small
    \centering
    \setlength\tabcolsep{4.2pt}
    \begin{tabular}{lcccccccc}
        \toprule
        \textbf{Domain} & \multicolumn{4}{c}{\textbf{Geometry}} & \multicolumn{3}{c}{\textbf{Table, Chart, and Figure}} & \textbf{Object} \\ 
        \cmidrule(lr){2-5} \cmidrule(lr){6-8} \cmidrule(lr){9-9}
        \textbf{Dataset} & Geos & GeoQA+ & Geometry3K & UniGeo & TabMWP & FigureQA & ChartQA & CLEVR \\ 
        \midrule
        \# Samples & 279 & 563 & 551 & 555 & 568 & 589 & 509 & 548 \\ 
        \bottomrule
    \end{tabular}
    \label{tab:data_summarize}
\end{table}

To complete these problems with long thought processes, we consider two approaches: either distilling from existing slow-thinking MLLMs or utilizing our fine-tuned MLLMs with textual long thought data. We assume that fine-tuning MLLMs with textual long thought data can effectively transform them into slow-thinking MLLMs, essentially engaging in a self-distillation process. For existing slow-thinking MLLMs, we select the recently released QVQ model, which demonstrates superior performance on several challenging benchmarks.


To generate the reasoning process, we use the commonly employed rollout method by randomly sampling responses from both the QVQ model and our own. We set a special format to parse the final answer from the response for each problem, retaining only those problems that the models can successfully solve within a reasonable number of rollouts. Intuitively, simpler problems require fewer rollouts to solve. We will further discuss the impact of problem difficulty on the fine-tuning performance of MLLMs in Section~\ref{sec:analysis}.

\subsubsection{Visual Long Thought Instruction Tuning}
\label{sec:visul_it}

When distilling the long thought data from QVQ (denoted by $D_{\text{QVQ}}$), the training process is straightforward: we fine-tune only the parameters of the LLM and the modality connector, as we do with the textual long thought data described in Section~\ref{sec:text-transfer}. Although the visual instruction data includes image information, our experimental results indicate that updating the visual encoder does not result in substantial performance improvement. 




As another alternative approach, we design a multi-stage tuning method for self-distillation. Specifically, we first fine-tune the selected MLLM (\ie Qwen2-VL-72B-Instruct) on the textual long thought instruction set $D_{\text{T}}$, obtaining model $\mathcal{M}_0$. Next, we use $\mathcal{M}_0$ to generate the visual long thought instruction set by self-distillation $D_{\text{SD}}$, which can be subsequently used for fine-tuning the original MLLM. 

In our experiments, our aim is to  investigate the effects of individual instruction datasets (\ie $D_{\text{T}}$, $D_{\text{SD}}$ and $D_{\text{QVQ}}$) and their combinations on the slow-thinking performance.  

\section{Experiments}
\subsection{Evaluation Setup}

To validate the effectiveness of our methods, we conduct experiments on four challenging benchmarks: MathVerse~\cite{zhang2025mathverse}, MathVision~\cite{wang2024measuring}, OlympiadBench~\cite{he2024olympiadbench}, and MMMU~\cite{yue2024mmmu}. MathVerse consists of 2,612 multi-subject math problems from diverse sources. MathVision comprises 3,040 high-quality mathematical problems sourced from established mathematics competitions. OlympiadBench
features 8,476 bilingual multimodal problems for Olympic-level mathematics and physics competitions.
MMMU encompasses 11,500 problems spanning 30 subjects and 183 subfields. To ensure a fair comparison, we conduct evaluations on the \emph{validation} set of MMMU and the {\emph{testmini}} set of MathVerse. Following VLMEvalKit~\cite{duan2024vlmevalkit}, we exclude the \emph{text-only} split from MathVerse and the theorem-proof parts from OlympiadBench. Among all the benchmarks, OlympiadBench is the most challenging, while MMMU demonstrates relatively lower difficulty levels and focuses more on comprehensive subject knowledge. 

We select Qwen2-VL-72B-Instruct~\cite{wang2024qwen2} as our base MLLM due to its strong multimodal capabilities. We fine-tune it with LLaMA-Factory~\cite{zheng2024llamafactory} and denote the resulting model as Virgo-72B. We then compare it with a range of models that are capable of conducting o1-like slow-thinking~(\ie OpenAI o1 and QVQ-72B-preview). We also include advanced general-purpose models~(\ie GPT-4o, Gemini-Pro, and Claude-3.5-Sonnet) for comparison. We also train Virgo-7B based on Qwen2-VL-7B-Instruct to further study the influence of model size.


\subsection{Main Results}
\label{sec:main_results}

\begin{table}[t]
    \centering
    \small
    \setlength\tabcolsep{2.8pt}
\caption{Performance comparison of top-tier MLLMs on four representative benchmarks. Here,  $D_{\text{T}}$ denotes the textual long thought data, and $D_{\text{SD}}$ and $D_{\text{QVQ}}$ denote the visual long thought data distilled by our model (the version fine-tuned by $D_{\text{T}}$) and QVQ, respectively.  
    The \textbf{bold} fonts denote the best performance among our training variants, while the \underline{underline} fonts denote the second-best performance. * Since QVQ has not released the evaluation code, we report the evaluation results reproduced by our team.} 
    \label{tab:main_res}
    \begin{tabular}{l|cc|c|c|c|c|c}
    \toprule
    \textbf{Model}& \multicolumn{2}{c|}{\textbf{Num. Data}} & \multirow{2}{*}{\textbf{MathVerse}}& \multirow{2}{*}{\textbf{MathVision}}& \multirow{2}{*}{\textbf{\makecell[c]{Olympiad\\Bench}}}& \multirow{2}{*}{\textbf{MMMU}} &\multirow{2}{*}{\textbf{Average}}\\ 
     & \textbf{Textual}& \textbf{Visual}& & & &  &\\
    \midrule
    GPT-4o & - & - & -& 30.4& 25.9& 69.1 &-\\
 Gemini-Pro& -& -& 35.3& 19.2& 4.2&65.8 &31.13\\
    Claude-3.5-Sonnet & - & - & -& 38.0& -& 70.4 &-\\
    OpenAI o1& - & - & -& -& -& 77.3 &-\\
    QVQ-72B-preview*& - & - & 41.5& 38.2& 27.9& 66.0 &43.40\\
 \midrule
 
 Qwen2-VL-72B-Instruct*& - & - & 41.3& 26.1& 11.2&64.5 &35.78\\
 Virgo-72B$_{D_\text{T}}$ & 5K& - & \textbf{48.4}& \textbf{38.8}& \textbf{29.9}& 64.6 &\textbf{45.43}\\
 Virgo-72B$_{D_\text{QVQ}}$& - & 6.6K & 37.6& 37.7  & 25.0 & 62.6 & 40.73\\
 Virgo-72B$_{D_\text{SD}}$& - & 7K& 47.4& 35.0& 27.2 & \textbf{65.8} &43.85\\
   Virgo-72B$_{D_\text{T} \cup D_\text{SD}}$& 5K& 7K& \underline{48.1}& {38.6}& 28.5& \underline{65.0} &\underline{45.05}\\
    \midrule
    
    Qwen2-VL-7B-Instruct& - & - & 24.6& 16.3& 5.3& \textbf{54.1} &25.08\\
    Virgo-7B$_{D_\text{T}}$& 5K& - & 32.2& \textbf{24.3}& 9.8& 47.1 &28.35\\
 Virgo-7B$_{D_\text{QVQ}}$& -& 6.6K&29.2 & 20.5 &  9.0 &48.3 &26.75\\
 Virgo-7B$_{D_\text{SD}}$& - & 7K&\textbf{37.5} & 23.1 & \textbf{10.3} &\underline{50.7} &\textbf{30.40}\\
 Virgo-7B$_{D_\text{T} \cup D_\text{SD}}$& 5K& 7K&\underline{36.7} & \underline{24.0} & \underline{10.2} &46.7 &\underline{29.40}\\
 \bottomrule
    \end{tabular}
\end{table}

In this section, we provide a comprehensive performance comparison of various methods on the selected evaluation benchmarks, as summarized in Table~\ref{tab:main_res}. The results include the performance of o1-like MLLMs, general-purpose MLLMs, and our approaches that extend the backbone model with different long thought instruction datasets. 


First, the slow-thinking reasoning ability can be effectively transferred through text-only reasoning data. As demonstrated in the second group of Table~\ref{tab:main_res}, after fine-tuning with only 5K textual long thought instructions, our model yields highly competitive results, approaching and even surpassing those of industry counterparts. For instance, our model achieves 38.4\% accuracy on MathVision and 29.3\% accuracy on OlympiadBench. However, another observation is that our model does not show significant improvement on the MMMU benchmark. 
To thoroughly analyze the performance limitations on MMMU, we further examine fine-grained performance by using the difficulty annotation of the test samples: easy, medium, and hard. As shown in Table~\ref{tab:difficulty}, our method lags behind QVQ in overall performance, with the disadvantage mainly concentrated in the easy and medium samples. For samples in the hard bin, our method achieves an accuracy of 54.70\%, compared to QVQ's 48.62\%. 
As we will discuss in Section~\ref{sec:case}, not all visual problems require complex reasoning processes, and enforcing a longer thought process might lead to performance degradation of MLLMs.

Secondly, synthesized visual instructions, whether obtained through distillation or self-distillation, do not significantly outperform textual reasoning instructions when fine-tuning the MLLM. Upon conducting a human review of the synthesized trajectories for visual questions, we find that many questions are not sufficiently complex and rely more on perception than reasoning, despite that we have carefully selected the data source and conducted a rigorous data filtering process to control the difficulty. Developing high-quality, complex visual instructions remains a challenging direction for future exploration.



Additionally, we conduct experiments on smaller MLLMs, specifically Qwen2-VL-7B-Instruct, as shown in the third group of Table~\ref{tab:main_res}. The performance trends observed with different reasoning instruction datasets show some deviations from those of the larger model, Qwen2-VL-72B-Instruct. Notably, Virgo-7B$_{D_{\text{SD}}}$ outperforms Virgo-7B$_{D_{\text{T}}}$, particularly on MathVerse and MMMU, suggesting that visual long-thinking instructions are more effective than textual instructions for small MLLM. Another difference is that after fine-tuning with long thought instructions, the performance on MMMU has substantially decreased. We speculate that a smaller model might be less capable of managing complex long thought processes, especially when applied to problems that do not necessitate complex reasoning (as MMMU appears to be simpler than the other three benchmarks). Incorporating visual instructions may alleviate this degradation.


\ignore{Beyond the results on slow-thinking reasoning ability transfer and visual adaptation, another intriguing question arises: "If high-quality visual instructions are available, would they outperform textual instructions?" To explore this, we conduct experiments on Qwen2-VL-7B-Instruct, using synthesized trajectories from Qwen2-VL-72B-Instruct as high-quality visual instructions. The results in the third group indicate that the model fine-tuned on textual instructions achieves comparable performance to the one fine-tuned on visual instructions. The only exception is that incorporating visual instructions is obviously beneficial for MathVerse. This is because many questions in MathVerse are in the same domain as our visual instructions.}

\subsection{Further Analysis}
\label{sec:analysis}
\begin{figure}[t]
    \centering    \includegraphics[width=0.75\textwidth]{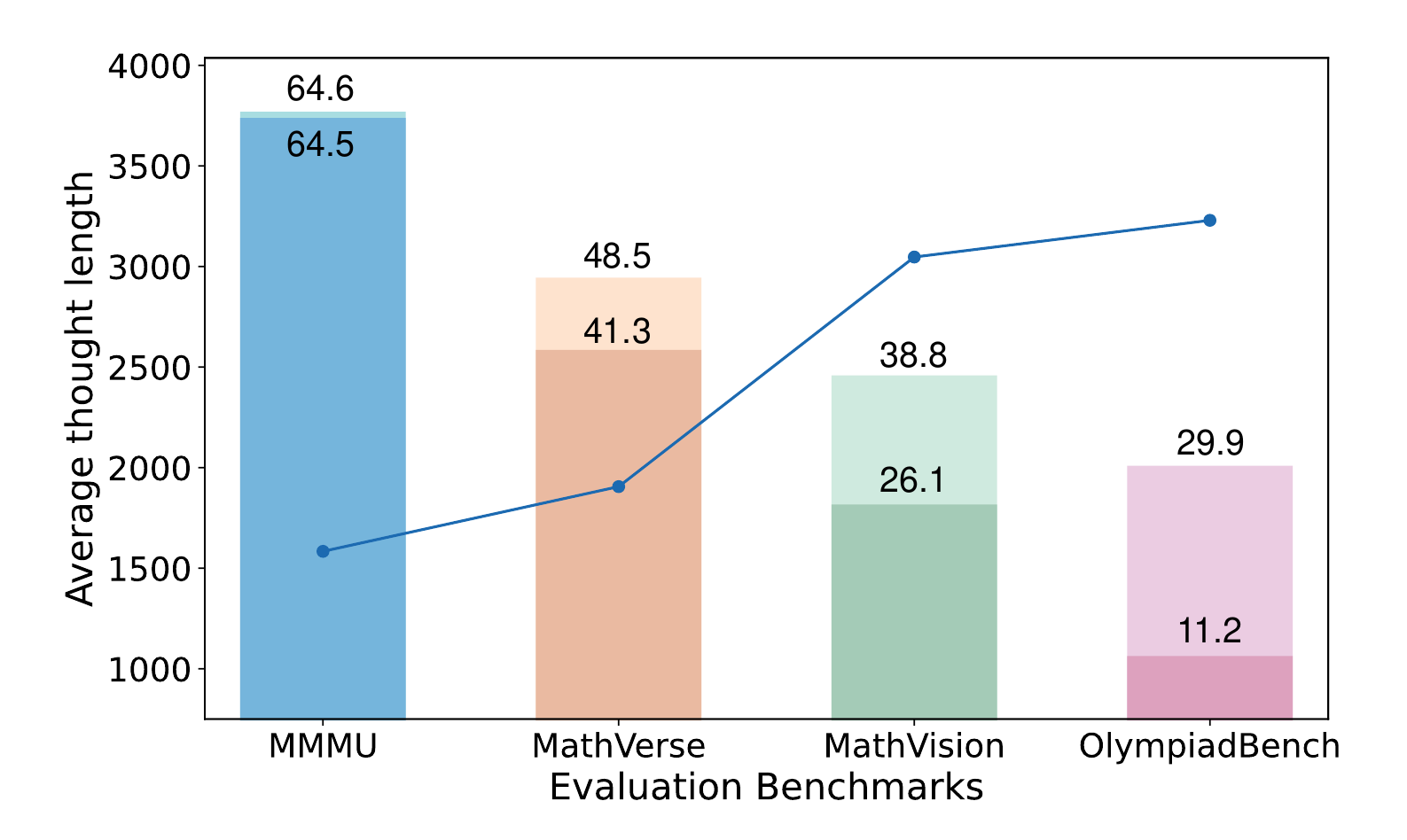}
  \caption{The relationship between the average thought length of each benchmark and the corresponding performance of both Virgo and Qwen2-VL-72B-Instruct. The ``average thought length'' is represented by the line, while ``performance'' is indicated by the bar. The bars in light color represent Vigor's performance, while the bars in dark color represent Qwen2-VL-72B-Instruct's performance. We observe that benchmarks with longer thought lengths generally correspond to greater performance improvements.} 
  \label{fig:response_length} 
\end{figure}

After presenting the overall performance analysis, we further investigate the detailed effects of long thought instruction data on visual reasoning. We present the major findings below.

\paragraph{Harder tasks benefit more from long thought reasoning.} 
We first examine how our approach impacts model performance across tasks of varying difficulty levels. Previous research\cite{min2024imitate} has indicated a correlation between the average length of responses generated by models and the complexity of the questions: longer responses generally accompany more complex or challenging questions. Building on this insight, we analyze the average length of responses produced by our model on evaluation benchmarks and visualize the corresponding model performance in Figure~\ref{fig:response_length}. The results indicate that benchmarks with longer response lengths, such as OlympiadBench, tend to be more difficult, as evidenced by their lower accuracy. Notably, our approach demonstrates substantial improvements on these challenging benchmarks, achieving absolute gains of 18.1\% and 12.4\% on OlympiadBench and MathVision, respectively. Conversely, we observe limited performance gains on the relatively easier benchmark, MMMU, which is characterized by shorter response lengths. 

\ignore{To perform a thorough analysis of performance on MMMU, we report the model performance on different splits of MMMU according to their difficulty levels in Table~\ref{tab:difficulty}. We observe that while our methods lag behind QVQ in overall performance, the discrepancy is mainly concentrated in the easy and medium splits. Significantly, on the hard split, our methods outperform QVQ, achieving an accuracy of 54.70 compared to QVQ's 48.62. This further demonstrates the superiority of our method on complex visual questions.}

\begin{table}[htbp]
    \centering
    \small
    \caption{Performance comparison on samples from different difficulty bins in MMMU.}\label{tab:difficulty}
      \begin{tabular}{l|cccc}
      \toprule
        \textbf{Model} & 
        \textbf{Easy}&
        \textbf{Medium}&
        \textbf{Hard}&\textbf{Overall}\\ 
        \midrule
      QVQ-72B-preview & \textbf{76.95} & \textbf{65.80} & 48.62   &\textbf{66.0}  \\ 
      Qwen2-VL-72B-Instruct & 74.58 & 62.26 & 50.28   &64.5  \\ 
      
      Virgo-72B$_{D_\text{T}}$ & 72.88 & \underline{62.97} & \underline{54.70}  &64.6 \\ 
      Virgo-72B$_{D_\text{T} \cup D_\text{SD}}$ & \underline{74.58} & 61.79 & \textbf{56.91}  &\underline{65.0} \\ 
      \bottomrule
      \end{tabular}
\end{table} 

\begin{figure}[t]
    \centering    \includegraphics[width=1.0\textwidth]{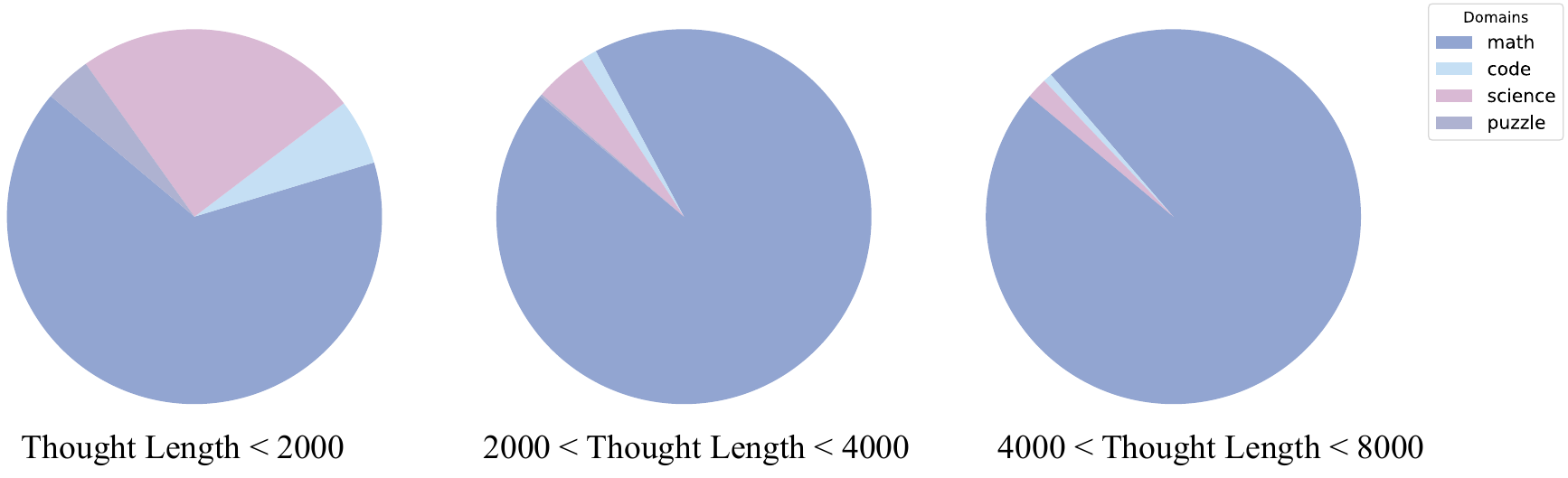}
  \caption{The domain distribution of textual long thought instructions.}
  \label{fig:domain_distribution} 
\end{figure}

\paragraph{Longer reasoning does not guarantee better results.} Since reasoning capacity is influenced by the difficulty of the instruction data, we compare fine-tuning performance across different difficulty levels. We use a simple method to determine instruction difficulty based on instruction length. Specifically, we train the model using textual long-thought instructions sampled from varying length ranges: \((0, 2000], (2000, 4000],\) and \((4000, 8000]\), and present the results in Table~\ref{tab:length}. 
The results indicate that increasing the length of reasoning in the training data from 2000 to 4000 tokens leads to performance improvements across all benchmarks. However, further increasing the length to 8000 tokens results in performance degradation on most benchmarks. To further examine the performance decrease associated with long instructions, we analyze the data composition of each length range and observe that the math domain dominates the long instruction data in the \((4000,8000]\)  range. These math problems may result in excessively long instructions compared to the actual required lengths for visual reasoning tasks; even the longest OlympiadBench examples have an average length below 4000, as shown in Figure~\ref{fig:response_length}.
\begin{table}[htbp]
 \setlength\tabcolsep{3.0pt}
    \centering
    \small
    \caption{Performance comparison by tuning with instruction datasets of varying thought length.}
      \begin{tabular}{l|ccccc}
      \toprule
        \textbf{Base Model} &  
        \textbf{Length}&
        \textbf{\makecell{MathVerse}} & 
        \textbf{\makecell{MathVision}} & \textbf{\makecell{OlympiadBench}} & \textbf{\makecell{MMMU}} \\ 
    \midrule
      \multirow{4}{*}{Qwen2-VL-7B-Instruct} & - & 24.0 & 15.6 & 5.3 & \textbf{54.1}\\ 
   & (0, 2000]& \underline{28.1}& 22.1& \underline{8.8}& 41.9\\
   & (2000, 4000]& \textbf{33.6}& \underline{24.4}& \textbf{9.1}& \underline{48.0}\\
   & (4000, 8000]& 24.8& \textbf{24.9}& 8.3& 39.8\\
      \bottomrule
      \end{tabular}
\label{tab:length}
\end{table}

\paragraph{Scaling textual instruction leads to improvement.} We further investigate the impact of scaling textual instructions on reasoning performance. The results, presented in Table~\ref{scaling}, demonstrate that increasing the number of textual instructions generally leads to performance improvements across most benchmarks. Specifically, increasing the instruction samples from 1K to 5K results in a 7.7\% average performance gain for both the 7B and 72B models on MathVision, while showing a modest 1.8\% performance gain on OlympiadBench. These observations suggest that while scaling textual instructions is generally effective, its impact varies across different tasks. Another finding is that textual instructions initially diminish the model's capacity on MMMU, but performance gradually recovers as more instructions are added.

\begin{table}[htbp]
 \setlength\tabcolsep{3.0pt}
    \centering
    \small
    \caption{The scaling effect of instruction data on the base model. }
      \begin{tabular}{l|ccccc}
      \toprule
        \textbf{Base Model} &  
        \textbf{Num.} &
        \textbf{\makecell{MathVerse}} & 
        \textbf{\makecell{MathVision}} & \textbf{\makecell{OlympiadBench}} & \textbf{\makecell{MMMU}} \\ 
  \midrule
      \multirow{4}{*}{Qwen2-VL-72B-Instruct} & 
      -  & 41.3 & 26.1 & 11.2 & 64.5   \\ 
     & 1K & 42.5 & 39.5 & 26.2 & 61.8   \\
     & 3K & 44.4 & \textbf{40.5} & 26.4 & 58.2   \\
     & 5K & \textbf{48.4} & 38.8 & \textbf{29.9} & \textbf{64.7}   \\
    \midrule
      \multirow{4}{*}{Qwen2-VL-7B-Instruct} & - & 24.0 & 15.6 & 5.3 & \textbf{54.1}\\ 
   & 1K & 22.5 & 23.7 & 8.6 & 42.8  \\
   & 3K & 30.2 & \textbf{24.9} & \textbf{9.6} & 44.6  \\
   & 5K  & \textbf{31.9} & 24.6 & 9.2 & 47.1  \\
      \bottomrule
      \end{tabular}
\label{scaling}
\end{table}



\paragraph{Difficulty of visual thought data has limited impacts on performance.} In Section~\ref{sec:visul_it}, we select visual problems from various domains and generate visual long thought instructions by distilling from QVQ and Virgo-72B$_{D_\text{T}}$. Our goal is to explore the impact of visual instructions with varying difficulty levels. Specifically, we first use Qwen2-VL-72B-Instruct, which has not been fine-tuned with long thought instructions, to generate responses for visual questions via greedy search. Questions that the base MLLM answers correctly are excluded, as they are considered relatively easy. For the remaining questions, Virgo-72B$_{D_\text{T}}$ performs multiple rollouts, generating five candidate trajectories per question. Based on the ratio of correct trajectories, we define two levels of difficulty: medium, for questions with 4 or 5 correct trajectories, and hard, for those with 2 or 3 correct trajectories. To investigate how question difficulty affects model performance, we also randomly sample some questions, regardless of whether the base MLLM can solve them, and synthesize trajectories based on these questions. This set is referred to as the ``random-level''. We combine 5K textual long thought instructions with each of the three splits (medium, hard, and random) to fine-tune Qwen2-VL-72B-Instruct and report the results in Table~\ref{tab:inst_difficulty}. The results show that visual instructions with different difficulty levels do not lead to significant performance differences. This suggests that advanced strategies for synthesizing visual long thought instructions are needed to enhance multi-modal slow-thinking reasoning capabilities.
\begin{table}[htbp]
 \setlength\tabcolsep{3.0pt}
    \centering
    \small
    \caption{Performance comparison of using self-distilled visual instruction of varying difficulty levels.}
      \begin{tabular}{l|ccccc}
      \toprule
        \textbf{Base Model} &  
        \textbf{Difficulty}&
        \textbf{\makecell{MathVerse}} & 
        \textbf{\makecell{MathVision}} & \textbf{\makecell{OlympiadBench}} & \textbf{\makecell{MMMU}} \\ 
    \midrule
      \multirow{4}[3]{*}{Qwen2-VL-72B-Instruct}& - & 41.3& 26.1& 11.2& 64.5\\ 
   & Medium& \textbf{48.1}& \underline{38.6}& 28.5& \textbf{65.0}\\
   & Hard& 47.4& \textbf{39.1}&\textbf{29.7}& 63.0\\
   & Random& \underline{47.9}& 38.5& \underline{29.3}& \underline{64.8}\\
      \bottomrule
      \end{tabular}
\label{tab:inst_difficulty}
\end{table}

\subsection{Case Study}
\label{sec:case}
In this section, we present several examples to demonstrate the advantages of slow-thinking reasoning in addressing complex multimodal problems. Additionally, we provide examples highlighting some of the negative impacts introduced by our approach.

\paragraph{Textual long thought instruction tuning elicits visual slow-thinking reasoning ability.} In Table~\ref{tab:case1}, the query requires evaluating the integral of a function and involves an image composed of three semi-circles. Qwen2-VL-72B-Instruct directly calculates the radius and center of each semi-circle individually but makes errors in determining their centers. In contrast, our model first describes the image in detail (highlighted in orange), then thoroughly reasons through the question, and finally arrives at the correct answer. Furthermore, the model can reflect on its reasoning process and attempt to verify its solution (highlighted in blue). This case demonstrates that long thought training enhances both the model's detailed captioning ability and its capacity for self-reflection, which are crucial for performing complex reasoning tasks.

\paragraph{Lack of reflection on perception causes reasoning to fail.} By examining several failure cases, we observe that Virgo fails to reflect on its perception results, which can cause the entire reasoning process to collapse. A representative case is illustrated in Table~\ref{tab:case2}, where Virgo mistakenly perceives the number of unemployed individuals with a ``high school diploma'' in September (highlighted in red). This leads to the incorrect conclusion that both August and September satisfy the problem's requirements. While Virgo recognizes the irrationality of the result and begins to reflect on its reasoning process (highlighted in blue), it does not question the validity of its perception. As a result, erroneous conclusions are repeatedly generated, leading to incorrect answers. This case highlights that slow-thinking MLLMs transferred from text-only instructions may have limited capacity for reflecting on perception. Future models should be designed with the ability to reflect on both perception results and reasoning processes.

\section{Conclusion}

In this paper, we have proposed a simple yet effective approach to implementing slow-thinking reasoning systems. Our core finding is that by simply fine-tuning MLLMs with textual long-form thought data, a capable MLLM can exhibit substantially enhanced slow-thinking capacities. We have conducted extensive experiments on four challenging benchmarks, and the results demonstrate that our approach achieves highly competitive performance when compared to industry reasoning systems. 
We have also examined the effect of different factors of textual instruction data, such as  data amount and  length distribution. 

We are aware that our current attempt at building multimodal slow-thinking systems is preliminary. For future work, we aim to expand the sources of challenging multimodal problems and design more principled methods to enhance this capacity.

\begin{table}[htbp]
    \centering
    \renewcommand{\arraystretch}{1.2} 
    \setlength{\tabcolsep}{8pt} 
    \begin{tabular}{p{14cm}}
        \centering
        \includegraphics[width=0.6\textwidth]{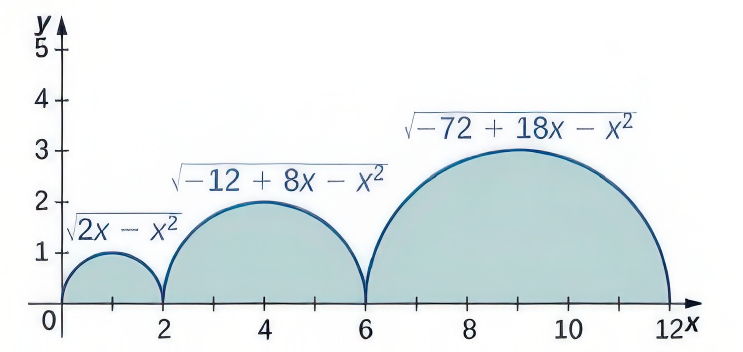}
        
        \begin{tcolorbox}[colback=gray!5, colframe=gray!40, sharp corners, width=\linewidth, boxrule=0.5mm]
        \small
        \textbf{Problem: }
        Evaluate the integral of the functions graphed using the formula for circles.
        
        \textbf{Ground Truth:} $7\pi$
        \end{tcolorbox}  
        \begin{tcolorbox}[colback=gray!5, colframe=gray!40, sharp corners, width=\linewidth, boxrule=0.5mm]
            \small
            \textbf{Qwen2-VL-72B-Instruct:}
            
To evaluate the integral of the functions graphed using the formula for circles, we need to identify the equations of the circles and then integrate them over the appropriate intervals.

1. **Identify the Equations of the Circles:**
   - The equation of a circle with center \((h, k)\) and radius \(r\) is given by:     $(x - h)^2 + (y - k)^2 = r^2$

2. **Determine the Centers and Radii:**
   - For the smallest circle: $2x - x^2 = 0 \implies x(2 - x) = 0 \implies x = 0 \text{ or } x = 2$ The center is at \((1, 1)\) and the radius is 1.

   - For the medium-sized circle:     $-12 + 8x - x^2 = 0 \implies -(x^2 - 8x + 12) = 0 \implies x^2 - 8x + 12 = 0$     Solving this quadratic equation:     $x = \frac{8 \pm \sqrt{64 - 48}}{2} = \frac{8 \pm 4}{2} = 4 \pm 2$

     So, the centers are at \((2, 2)\) and \((6, 2)\). The radii are 2.

   - For the largest circle:     $-72 + 18x - x^2 = 0 \implies -(x^2 - 18x + 72) = 0 \implies x^2 - 18x + 72 = 0$

     Solving this quadratic equation:     $x = \frac{18 \pm \sqrt{324 - 288}}{2} = \frac{18 \pm 6}{2} = 9 \pm 3$

     So, the centers are at \((6, 3)\) and \((12, 3)\). The radii are 3.

3. **Integrate the Functions:**
   - For each circle, we integrate the function over the interval where it is non-zero.
   - The integrals will be over the intervals determined by the x-coordinates of the centers and the radii.

4. **Final Answer:**
   The integral of the functions graphed using the formula for circles is:
   
   $$\boxed{18\pi}$$

        \end{tcolorbox}
        
        \begin{tcolorbox}[colback=gray!5, colframe=gray!40, sharp corners, width=\linewidth, boxrule=0.5mm]
        \small
        \textbf{Virgo-72B (ours): }
        \\
        
        \textit{<|begin\_of\_thought|>}\\

Alright, I've got this graph here with three semicircles plotted. \textcolor{orange}{The x-axis ranges from 0 to 12, and the y-axis goes up to 5. Each semicircle is labeled with an expression involving x. The smallest one on the left is labeled $\sqrt{1 - (x - 1)^2}$, the middle one is $\sqrt{-12 + 8x - x^2}$, and the largest one on the right is $\sqrt{-72 + 18x - x^2}$. All of them are filled with a light blue color, and there are arrows pointing upwards from the x-axis to the top of each semicircle, indicating the height at different points.}

Let's start with the smallest one: $\sqrt{(2x - x^2)}$. I can factor out $-x^2$ from the expression: $\sqrt{(-x^2 + 2x)} = \sqrt{(-x^2 + 2x - 1 + 1)} = \sqrt{(-(x - 1)^2 + 1)}$. So, this simplifies to $\sqrt{(1 - (x - 1)^2)}$. Comparing this to the standard form, this seems like the upper half of a circle centered at (1, 0) with radius 1.

Next, the middle one: $\sqrt{(-12 + 8x - x^2)}$...

Now that I've identified the equations of these semicircles, I need to evaluate the integral of the functions graphed using the formula for circles. I'm assuming that means finding the area under each semicircle from their respective starting points to their endpoints on the x-axis.

Since each of these is a semicircle, the area under each curve would be half the area of the full circle. The area of a full circle is $\pi r^2$, so for a semicircle, it's $\frac{1}{2}\pi r^2$.

\end{tcolorbox}
\end{tabular}
\caption{An example that Virgo-72B outperforms Qwen2-VL-72B-Instruct (part 1).}
\label{tab:case1}
\end{table}

\begin{table}[htbp]
    \ContinuedFloat
    \begin{tabular}{p{14cm}}
\begin{tcolorbox}[colback=gray!5, colframe=gray!40, sharp corners, width=\linewidth, boxrule=0.5mm]
\small

Let's calculate the area for each semicircle:

1. Smallest semicircle: radius $r = 1$, so area = $\frac{1}{2}\pi (1)^2 = \frac{\pi}{2}$.

2. Middle semicircle: 

......

So, the total area under all three semicircles is $7\pi$.

\textcolor{blue}{Wait a minute, but the question says "evaluate the integral of the functions graphed using the formula for circles." Does that mean I need to integrate each function separately and then sum them up?}

\textcolor{blue}{Let me double-check. If I integrate each function from their starting point to their endpoint, I should get the area under each curve.}

......

Given that, I think my initial approach is correct: the integral of each function over its domain gives the area under the semicircle, which is half the area of the full circle.

So, the final answer should be $7\pi$.

\textit{<|end\_of\_thought|>}\\

\textit{<|begin\_of\_solution|>}\\

To evaluate the integral of the functions graphed using the formula for circles, we first identify the equations of the semicircles from the given expressions:

1. The smallest semicircle is represented by $\sqrt{2x - x^2}$, which simplifies to $\sqrt{1 - (x - 1)^2}$. This is the upper half of a circle centered at $(1, 0)$ with radius 1.

......

Thus, the final answer is:

$$\boxed{7\pi}$$\\
\textit{<|end\_of\_solution|>}\\
        \end{tcolorbox}
    \end{tabular}
    \caption{An example that Virgo-72B outperforms Qwen2-VL-72B-Instruct (part 2)}
    
\end{table}
\newpage

\begin{table}[htbp]
    \centering
    \renewcommand{\arraystretch}{1.2} 
    \setlength{\tabcolsep}{8pt} 
    \begin{tabular}{p{14cm}}
        \centering
        \includegraphics[width=0.5\textwidth]{fig/validation_Manage_8.pdf}
        
        \begin{tcolorbox}[colback=gray!5, colframe=gray!40, sharp corners, width=\linewidth, boxrule=0.5mm]
        \small
        \textbf{Problem: }Graph 1 shows the number of unemployed (in thousands) over a five-month period, organized by level of education completed. The difference between the number of unemployed high school graduates and the number of unemployed workers without a high school diploma was smallest in \underline{ }\underline{ }\underline{ }\underline{ }\underline{ }\underline{ }\underline{ }\underline{ }\underline{ }\underline{ }\underline{ }.

A. June

B. July

C. August

D. September

E. October
        
        \textbf{Ground Truth:} D. September
        \end{tcolorbox}  
        \begin{tcolorbox}[colback=gray!5, colframe=gray!40, sharp corners, width=\linewidth, boxrule=0.5mm]
            \small
            \textbf{Qwen2-VL-72B-Instrct:}

            D. September
            
        \end{tcolorbox}
        
        \begin{tcolorbox}[colback=gray!5, colframe=gray!40, sharp corners, width=\linewidth, boxrule=0.5mm]
        \small
        \textbf{Virgo-72B (ours): }
        \\
        
        \textit{<|begin\_of\_thought|>}\\

Alright, so I've got this graph here that shows the number of unemployed people, in thousands, over a five-month period, from June to October. It breaks down the unemployment numbers by different levels of education completed. There are four categories: ``Less than a high school diploma'', ``High school diploma, no college'', ``Some college'', and ``Bachelor's degree and higher''. Each category has a line representing the number of unemployed workers each month.\\

I need to find out in which month the difference between the number of unemployed high school graduates and the number of unemployed workers without a high school diploma was the smallest. So, I need to look at the lines for ``High school diploma, no college'' and ``Less than a high school diploma'' and see when the gap between them is the smallest.\\

...... \\

In August, high school graduates are around 8.5 thousand, and those without a high school diploma are about 11.5 thousand. The difference is about 3 thousand.\\

\textcolor{red}{In September, high school graduates are around 8 thousand}, and those without a high school diploma are about 11 thousand. The difference is about 3 thousand.\\

\end{tcolorbox}
\end{tabular}
\caption{An example that Virgo-72B fails to reflect on perception results (part 1).}
\label{tab:case2}
\end{table}

\begin{table}[htbp]
    \ContinuedFloat
    \begin{tabular}{p{14cm}}
\begin{tcolorbox}[colback=gray!5, colframe=gray!40, sharp corners, width=\linewidth, boxrule=0.5mm]
\small

In October, high school graduates are around 8 thousand, and those without a high school diploma are about 12 thousand. The difference is about 4 thousand.\\

So, from this, it seems like in August and September, the difference is the smallest, about 3 thousand. 

......\\

In August, high school graduates are around 8.5 thousand, and those without a high school diploma are around 11.5 thousand. The difference is 3 thousand.\\

\textcolor{blue}{In September, high school graduates are around 8 thousand, and those without a high school diploma are around 11 thousand. The difference is 3 thousand.}\\

In October, high school graduates are around 8 thousand, and those without a high school diploma are around 12 thousand. The difference is 4 thousand.\\

So, the smallest difference is in August and September, both with a 3-thousand difference.\\

But I need to make sure if there's only one answer or multiple. The question says ``was smallest in \underline{ }\underline{ }\underline{ }\underline{ }\underline{ }\underline{ }\underline{ }\underline{ }\underline{ }\underline{ }\underline{ }\underline{ },'' which might imply a single answer, but looking back, it could be in both August and September.\\

Maybe I should consider if there's a tie for the smallest difference. \textcolor{blue}{Looking back, both August and September have the same difference of 3 thousand}, while the other months have larger differences. Therefore, the difference was smallest in both August and September. But since the question might expect a single answer, perhaps I need to choose the earliest month, which would be August. Alternatively, maybe both options are acceptable. But in any case, the smallest difference occurred in August and September.

\textit{<|end\_of\_thought|>}\\

\textit{<|begin\_of\_solution|>}\\

To determine the month with the smallest difference between the number of unemployed high school graduates and the number of unemployed workers without a high school diploma, we analyze the graph provided.\\

- In June, the difference is approximately 4 thousand.\\

- In July, the difference is approximately 3.5 thousand.\\

- In August, the difference is approximately 3 thousand.\\

- In September, the difference is approximately 3 thousand.\\

- In October, the difference is approximately 4 thousand.\\

The smallest difference occurs in both August and September, with a difference of 3 thousand each.\\

Thus, the correct answer is:\\

$$\boxed{C}$$

\textit{<|end\_of\_solution|>}

        \end{tcolorbox}
    \end{tabular}
    \caption{An example that Virgo-72B fails to reflect on perception results (part 2).}
\end{table}

\clearpage

\bibliographystyle{unsrt}
\bibliography{ref.bib}

\begin{thebibliography}{10}

\bibitem{o1}
OpenAI.
\newblock Learning to reason with large language models, September 2024.

\bibitem{r1}
DeepSeek Team.
\newblock Deepseek-r1-lite-preview is now live: unleashing supercharged reasoning power!, November 2024.

\bibitem{qwq}
Qwen Team.
\newblock Qwq: Reflect deeply on the boundaries of the unknown, November 2024.

\bibitem{zhao2023survey}
Wayne~Xin Zhao, Kun Zhou, Junyi Li, Tianyi Tang, Xiaolei Wang, Yupeng Hou, Yingqian Min, Beichen Zhang, Junjie Zhang, Zican Dong, Yifan Du, Chen Yang, Yushuo Chen, Zhipeng Chen, Jinhao Jiang, Ruiyang Ren, Yifan Li, Xinyu Tang, Zikang Liu, Peiyu Liu, Jian{-}Yun Nie, and Ji{-}Rong Wen.
\newblock A survey of large language models.
\newblock {\em CoRR}, abs/2303.18223, 2023.

\bibitem{wu2024comparative}
Siwei Wu, Zhongyuan Peng, Xinrun Du, Tuney Zheng, Minghao Liu, Jialong Wu, Jiachen Ma, Yizhi Li, Jian Yang, Wangchunshu Zhou, Qunshu Lin, Junbo Zhao, Zhaoxiang Zhang, Wenhao Huang, Ge~Zhang, Chenghua Lin, and Jiaheng Liu.
\newblock A comparative study on reasoning patterns of openai's o1 model.
\newblock {\em CoRR}, abs/2410.13639, 2024.

\bibitem{zhong2024evaluation}
Tianyang Zhong, Zhengliang Liu, Yi~Pan, Yutong Zhang, Yifan Zhou, Shizhe Liang, Zihao Wu, Yanjun Lyu, Peng Shu, Xiaowei Yu, Chao Cao, Hanqi Jiang, Hanxu Chen, Yiwei Li, Junhao Chen, Huawen Hu, Yihen Liu, Huaqin Zhao, Shaochen Xu, Haixing Dai, Lin Zhao, Ruidong Zhang, Wei Zhao, Zhenyuan Yang, Jingyuan Chen, Peilong Wang, Wei Ruan, Hui Wang, Huan Zhao, Jing Zhang, Yiming Ren, Shihuan Qin, Tong Chen, Jiaxi Li, Arif~Hassan Zidan, Afrar Jahin, Minheng Chen, Sichen Xia, Jason Holmes, Yan Zhuang, Jiaqi Wang, Bochen Xu, Weiran Xia, Jichao Yu, Kaibo Tang, Yaxuan Yang, Bolun Sun, Tao Yang, Guoyu Lu, Xianqiao Wang, Lilong Chai, He~Li, Jin Lu, Lichao Sun, Xin Zhang, Bao Ge, Xintao Hu, Lian Zhang, Hua Zhou, Lu~Zhang, Shu Zhang, Ninghao Liu, Bei Jiang, Linglong Kong, Zhen Xiang, Yudan Ren, Jun Liu, Xi~Jiang, Yu~Bao, Wei Zhang, Xiang Li, Gang Li, Wei Liu, Dinggang Shen, Andrea Sikora, Xiaoming Zhai, Dajiang Zhu, and Tianming Liu.
\newblock Evaluation of openai o1: Opportunities and challenges of {AGI}.
\newblock {\em CoRR}, abs/2409.18486, 2024.

\bibitem{openai2024reasoning}
OpenAI.
\newblock Learning to reason with llms, 2024.

\bibitem{qvq}
Qwen Team.
\newblock Qvq: To see the world with wisdom, December 2024.

\bibitem{jiang2024technical}
Jinhao Jiang, Zhipeng Chen, Yingqian Min, Jie Chen, Xiaoxue Cheng, Jiapeng Wang, Yiru Tang, Haoxiang Sun, Jia Deng, Wayne~Xin Zhao, et~al.
\newblock Technical report: Enhancing llm reasoning with reward-guided tree search.
\newblock {\em CoRR}, abs/2411.11694, 2024.

\bibitem{zhang2024llamaberry}
Di~Zhang, Jianbo Wu, Jingdi Lei, Tong Che, Jiatong Li, Tong Xie, Xiaoshui Huang, Shufei Zhang, Marco Pavone, Yuqiang Li, Wanli Ouyang, and Dongzhan Zhou.
\newblock Llama-berry: Pairwise optimization for o1-like olympiad-level mathematical reasoning.
\newblock {\em CoRR}, abs/2410.02884, 2024.

\bibitem{zhang2024o1coder}
Yuxiang Zhang, Shangxi Wu, Yuqi Yang, Jiangming Shu, Jinlin Xiao, Chao Kong, and Jitao Sang.
\newblock o1-coder: an o1 replication for coding.
\newblock {\em CoRR}, abs/2412.00154, 2024.

\bibitem{qin2024o1replicationjourneystrategic}
Yiwei Qin, Xuefeng Li, Haoyang Zou, Yixiu Liu, Shijie Xia, Zhen Huang, Yixin Ye, Weizhe Yuan, Hector Liu, Yuanzhi Li, and Pengfei Liu.
\newblock O1 replication journey: A strategic progress report -- part 1.
\newblock {\em CoRR}, 2024.

\bibitem{zhao2024marco}
Yu~Zhao, Huifeng Yin, Bo~Zeng, Hao Wang, Tianqi Shi, Chenyang Lyu, Longyue Wang, Weihua Luo, and Kaifu Zhang.
\newblock Marco-o1: Towards open reasoning models for open-ended solutions.
\newblock {\em CoRR}, abs/2411.14405, 2024.

\bibitem{li2024multimodal}
Chunyuan Li, Zhe Gan, Zhengyuan Yang, Jianwei Yang, Linjie Li, Lijuan Wang, Jianfeng Gao, et~al.
\newblock Multimodal foundation models: From specialists to general-purpose assistants.
\newblock {\em Foundations and Trends{\textregistered} in Computer Graphics and Vision}, 16(1-2):1--214, 2024.

\bibitem{yang2023mmreact}
Zhengyuan Yang, Linjie Li, Jianfeng Wang, Kevin Lin, Ehsan Azarnasab, Faisal Ahmed, Zicheng Liu, Ce~Liu, Michael Zeng, and Lijuan Wang.
\newblock {MM-REACT:} prompting chatgpt for multimodal reasoning and action.
\newblock {\em CoRR}, abs/2303.11381, 2023.

\bibitem{li2024temporal}
Lei Li, Yuanxin Liu, Linli Yao, Peiyuan Zhang, Chenxin An, Lean Wang, Xu~Sun, Lingpeng Kong, and Qi~Liu.
\newblock Temporal reasoning transfer from text to video.
\newblock {\em arXiv preprint arXiv:2410.06166}, 2024.

\bibitem{zhang2024long}
Peiyuan Zhang, Kaichen Zhang, Bo~Li, Guangtao Zeng, Jingkang Yang, Yuanhan Zhang, Ziyue Wang, Haoran Tan, Chunyuan Li, and Ziwei Liu.
\newblock Long context transfer from language to vision.
\newblock {\em arXiv preprint arXiv:2406.16852}, 2024.

\bibitem{min2024imitate}
Yingqian Min, Zhipeng Chen, Jinhao Jiang, Jie Chen, Jia Deng, Yiwen Hu, Yiru Tang, Jiapeng Wang, Xiaoxue Cheng, Huatong Song, et~al.
\newblock Imitate, explore, and self-improve: A reproduction report on slow-thinking reasoning systems.
\newblock {\em arXiv preprint arXiv:2412.09413}, 2024.

\bibitem{wang2024qwen2}
Peng Wang, Shuai Bai, Sinan Tan, Shijie Wang, Zhihao Fan, Jinze Bai, Keqin Chen, Xuejing Liu, Jialin Wang, Wenbin Ge, et~al.
\newblock Qwen2-vl: Enhancing vision-language model's perception of the world at any resolution.
\newblock {\em arXiv preprint arXiv:2409.12191}, 2024.

\bibitem{rafailov2024direct}
Rafael Rafailov, Archit Sharma, Eric Mitchell, Christopher~D. Manning, Stefano Ermon, and Chelsea Finn.
\newblock Direct preference optimization: Your language model is secretly a reward model.
\newblock In {\em NeurIPS}, 2023.

\bibitem{Ouyang2022instruct}
Long Ouyang, Jeffrey Wu, Xu~Jiang, Diogo Almeida, Carroll~L. Wainwright, Pamela Mishkin, Chong Zhang, Sandhini Agarwal, Katarina Slama, Alex Ray, John Schulman, Jacob Hilton, Fraser Kelton, Luke Miller, Maddie Simens, Amanda Askell, Peter Welinder, Paul~F. Christiano, Jan Leike, and Ryan Lowe.
\newblock Training language models to follow instructions with human feedback.
\newblock In {\em NeurIPS}, 2022.

\bibitem{seo2015solving}
Minjoon Seo, Hannaneh Hajishirzi, Ali Farhadi, Oren Etzioni, and Clint Malcolm.
\newblock Solving geometry problems: Combining text and diagram interpretation.
\newblock In {\em Proceedings of the 2015 conference on empirical methods in natural language processing}, pages 1466--1476, 2015.

\bibitem{chen2021geoqa}
Jiaqi Chen, Jianheng Tang, Jinghui Qin, Xiaodan Liang, Lingbo Liu, Eric~P Xing, and Liang Lin.
\newblock Geoqa: A geometric question answering benchmark towards multimodal numerical reasoning.
\newblock {\em arXiv preprint arXiv:2105.14517}, 2021.

\bibitem{lu2021inter}
Pan Lu, Ran Gong, Shibiao Jiang, Liang Qiu, Siyuan Huang, Xiaodan Liang, and Song-Chun Zhu.
\newblock Inter-gps: Interpretable geometry problem solving with formal language and symbolic reasoning.
\newblock {\em arXiv preprint arXiv:2105.04165}, 2021.

\bibitem{chen2022unigeo}
Jiaqi Chen, Tong Li, Jinghui Qin, Pan Lu, Liang Lin, Chongyu Chen, and Xiaodan Liang.
\newblock Unigeo: Unifying geometry logical reasoning via reformulating mathematical expression.
\newblock {\em arXiv preprint arXiv:2212.02746}, 2022.

\bibitem{lu2022dynamic}
Pan Lu, Liang Qiu, Kai-Wei Chang, Ying~Nian Wu, Song-Chun Zhu, Tanmay Rajpurohit, Peter Clark, and Ashwin Kalyan.
\newblock Dynamic prompt learning via policy gradient for semi-structured mathematical reasoning.
\newblock {\em arXiv preprint arXiv:2209.14610}, 2022.

\bibitem{kahou2017figureqa}
Samira~Ebrahimi Kahou, Vincent Michalski, Adam Atkinson, {\'A}kos K{\'a}d{\'a}r, Adam Trischler, and Yoshua Bengio.
\newblock Figureqa: An annotated figure dataset for visual reasoning.
\newblock {\em arXiv preprint arXiv:1710.07300}, 2017.

\bibitem{masry2022chartqa}
Ahmed Masry, Do~Xuan Long, Jia~Qing Tan, Shafiq Joty, and Enamul Hoque.
\newblock Chartqa: A benchmark for question answering about charts with visual and logical reasoning.
\newblock {\em arXiv preprint arXiv:2203.10244}, 2022.

\bibitem{johnson2017clevr}
Justin Johnson, Bharath Hariharan, Laurens Van Der~Maaten, Li~Fei-Fei, C~Lawrence~Zitnick, and Ross Girshick.
\newblock Clevr: A diagnostic dataset for compositional language and elementary visual reasoning.
\newblock In {\em Proceedings of the IEEE conference on computer vision and pattern recognition}, pages 2901--2910, 2017.

\bibitem{li2024llava}
Bo~Li, Yuanhan Zhang, Dong Guo, Renrui Zhang, Feng Li, Hao Zhang, Kaichen Zhang, Peiyuan Zhang, Yanwei Li, Ziwei Liu, et~al.
\newblock Llava-onevision: Easy visual task transfer.
\newblock {\em arXiv preprint arXiv:2408.03326}, 2024.

\bibitem{zhang2025mathverse}
Renrui Zhang, Dongzhi Jiang, Yichi Zhang, Haokun Lin, Ziyu Guo, Pengshuo Qiu, Aojun Zhou, Pan Lu, Kai-Wei Chang, Yu~Qiao, et~al.
\newblock Mathverse: Does your multi-modal llm truly see the diagrams in visual math problems?
\newblock In {\em European Conference on Computer Vision}, pages 169--186. Springer, 2025.

\bibitem{wang2024measuring}
Ke~Wang, Junting Pan, Weikang Shi, Zimu Lu, Mingjie Zhan, and Hongsheng Li.
\newblock Measuring multimodal mathematical reasoning with math-vision dataset.
\newblock {\em arXiv preprint arXiv:2402.14804}, 2024.

\bibitem{he2024olympiadbench}
Chaoqun He, Renjie Luo, Yuzhuo Bai, Shengding Hu, Zhen~Leng Thai, Junhao Shen, Jinyi Hu, Xu~Han, Yujie Huang, Yuxiang Zhang, et~al.
\newblock Olympiadbench: A challenging benchmark for promoting agi with olympiad-level bilingual multimodal scientific problems.
\newblock {\em arXiv preprint arXiv:2402.14008}, 2024.

\bibitem{yue2024mmmu}
Xiang Yue, Yuansheng Ni, Kai Zhang, Tianyu Zheng, Ruoqi Liu, Ge~Zhang, Samuel Stevens, Dongfu Jiang, Weiming Ren, Yuxuan Sun, et~al.
\newblock Mmmu: A massive multi-discipline multimodal understanding and reasoning benchmark for expert agi.
\newblock In {\em Proceedings of the IEEE/CVF Conference on Computer Vision and Pattern Recognition}, pages 9556--9567, 2024.

\bibitem{duan2024vlmevalkit}
Haodong Duan, Junming Yang, Yuxuan Qiao, Xinyu Fang, Lin Chen, Yuan Liu, Xiaoyi Dong, Yuhang Zang, Pan Zhang, Jiaqi Wang, et~al.
\newblock Vlmevalkit: An open-source toolkit for evaluating large multi-modality models.
\newblock In {\em Proceedings of the 32nd ACM international conference on multimedia}, pages 11198--11201, 2024.

\bibitem{zheng2024llamafactory}
Yaowei Zheng, Richong Zhang, Junhao Zhang, Yanhan Ye, Zheyan Luo, Zhangchi Feng, and Yongqiang Ma.
\newblock Llamafactory: Unified efficient fine-tuning of 100+ language models.
\newblock In {\em Proceedings of the 62nd Annual Meeting of the Association for Computational Linguistics (Volume 3: System Demonstrations)}, Bangkok, Thailand, 2024. Association for Computational Linguistics.

\end{thebibliography}

\end{document}